\pdfoutput=1

\documentclass[11pt]{article}

\usepackage[]{acl}
\usepackage{times}
\usepackage{latexsym}

\usepackage[T1]{fontenc}

\usepackage[utf8]{inputenc}

\usepackage{microtype}
\usepackage{graphicx}
\usepackage{multirow}
\usepackage{subfigure}
\usepackage{amsmath, amssymb, mathtools}

\title{Domain-Oriented Prefix-Tuning: Towards Efficient and Generalizable Fine-tuning for Zero-Shot Dialogue Summarization}

\author{Lulu Zhao$^{1*}$, {Fujia Zheng$^{1}$\thanks{\ \ \ The first two authors contribute equally.}}, Weihao Zeng$^{1}$, Keqing He$^{2}$, Weiran Xu$^{1}$\thanks{\ \ \  Weiran Xu is the corresponding author.}, 
\\{\bf Huixing Jiang$^{2}$}, {\bf Wei Wu$^{2}$}, {\bf Yanan Wu$^{1}$} \\
 $^1$Pattern Recognition \& Intelligent System Laboratory \\
  $^1$Beijing University of Posts and Telecommunications, Beijing, China\\
$^{2}$Meituan Group, Beijing, China\\
  \texttt{\{zhaoll,fujia\_zheng,ZengWH,xuweiran\}@bupt.edu.cn}\\
   \texttt{\{kqin\}@bupt.cn},
  \texttt{\{jhx\_bupt\}@163.com},
  \texttt{\{wuwei19850318\}@gmail.com}
  }

\begin{document}
\maketitle
\begin{abstract}
The most advanced abstractive dialogue summarizers lack generalization ability on new domains and the existing researches for domain adaptation in summarization generally rely on large-scale pre-trainings. To explore the lightweight fine-tuning methods for domain adaptation of dialogue summarization, in this paper, we propose an efficient and generalizable \textbf{D}omain-\textbf{O}riented \textbf{P}refix-tuning model, which utilizes a domain word initialized prefix module to alleviate domain entanglement and adopts discrete prompts to guide the model to focus on key contents of dialogues and enhance model generalization. We conduct zero-shot experiments and build domain adaptation benchmarks on two multi-domain dialogue summarization datasets, TODSum and QMSum. Adequate experiments and qualitative analysis prove the effectiveness of our methods.
\end{abstract}

\section{Introduction}

Abstractive dialogue summarization task aims to distill the most critical information in a conversation to produce a concise version, involving chit-chat \cite{Gliwa2019SAMSumCA,chen-yang-2020-multi}, meeting\cite{zhong-etal-2021-qmsum}, customer service \cite{Liu2019AutomaticDS,Zou2021TopicOrientedSD}, and task-oriented dialogue scenarios \cite{zhao2021todsum}. Compared to the single-speaker texts, summarizing a dialogue presents a unique set of challenges, such as unstructured expressions and information sparsity issues. Recently, large-scale generative pre-trained models \cite{Lewis2020BARTDS,liu2019text} have promoted the development of abstractive dialogue summarization but they all require extensive human-annotated golden summaries, which makes them not scalable to new domains where only few/no labeled data is available. Considering that real-world applications often face the problem of data in the new domain, it is vital to develop low-resource dialogue summarization models for the target domain by leveraging limited annotated data of source domains.

\begin{figure}[t]
\centering
\resizebox{0.47\textwidth}{!}{
\includegraphics[scale=0.65]{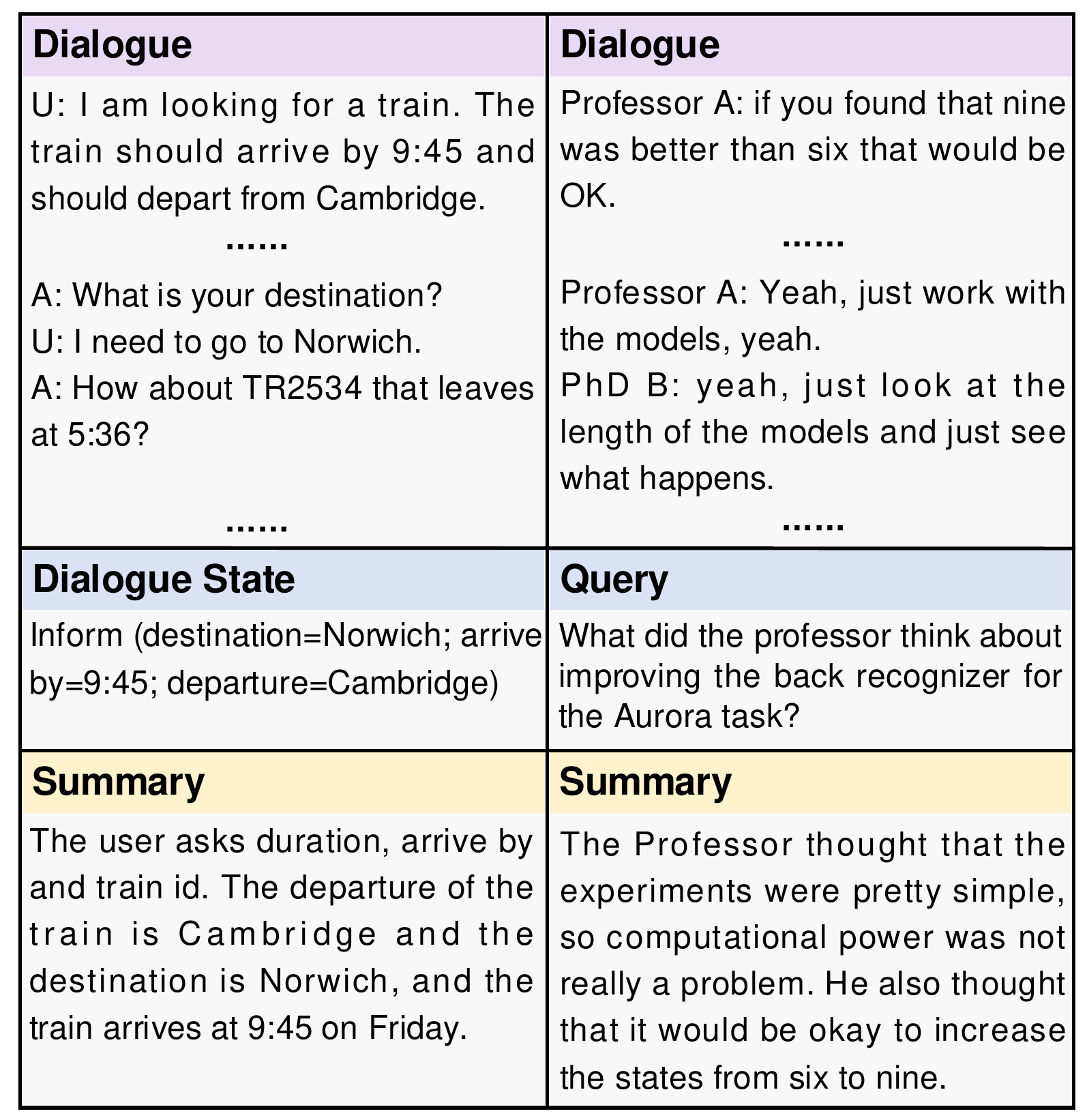}}
\caption{An example of TODSum dataset \cite{zhao2021todsum} with dialogue state and an example of QMSum dataset \cite{zhong-etal-2021-qmsum} with query. Note that dialogue state and query are existing characteristics of these two datasets, respectively.}
\label{fig:intro}
\vspace{-0.7cm}
\end{figure} 


Therefore, we try to explore the efficient domain adaptation of dialogue summarization models from the source domain $D_{s}$ to the target domain $D_{t}$, where $D_{s}$ only has limited annotated summaries and $D_{t}$ has few/no labeled data. There exist some domain adaptation approaches that focus on continual pre-trainings using some large domain/task-related corpora. \citet{yu-etal-2021-adaptsum} added multiple pre-training stages both on source domains and target domains. Further, \citet{zou-etal-2021-low} decomposed the pre-training into three procedures, i.e., the pre-training of encoder, decoder, and the combined encoder-decoder model. \citet{fabbri-etal-2021-improving} constructed pseudo summaries based on external Wikipedia data to simulate characteristics of target dataset. Note that all these methods require time-consuming pre-trainings or large-scale external corpora. They only focus on the heavy pre-training stage rather than the lightweight fine-tuning, which makes it labor-expensive and environmentally unfriendly \cite{Schwartz2020GreenA} to practical applications.

Different from existing works that adopt general pre-trainings on the large-scale external corpora, we focus on exploring efficient fine-tuning methods specifically targeted at domain adaptation for the dialogue summarization task. We consider the following key principles while designing our methods: (1) \textbf{Efficiency}: We do not use any external data or pre-training and aim at leveraging efficient fine-tuning mechanisms based on existing summarization models. (2) \textbf{Domain Disentanglement}: Traditional models often memorize excessive knowledge from $D_s$ and generate wrong summaries containing specific domain words in $D_s$. We aim to disentangle shared domain knowledge from $D_s$. (3) \textbf{Generalization}: Models often learn specific features of the source domain, making it difficult to generalize in a new domain \cite{pmlr-v97-peng19b}. For example, models learn the surface language style specific to $D_s$ rather than adapting the way of saliency estimation and summary generation to $D_t$. We encourage the summarizer to only focus on generic key contents rather than domain-specific attributes.

To be consistent with above principles, we propose a lightweight and efficient \textbf{D}omain-\textbf{O}riented \textbf{P}refix-tuning method, DOP, for domain adaptation of dialogue summarization. For efficiency, we focus on fine-tuning summarization models instead of performing pre-trainings like existing works, which reduces expensive and time-consuming computation. For domain disentanglement, we design a domain-oriented prefix module, which contains a novel prompt initialization mechanism. Concretely, we use domain words extracted by unsupervised LDA \cite{NIPS2010_3902} to initialize continuous prompt vectors and fit the outputs of MLP and pre-computed BART to obtain initial parameters and representations of the prefix module. We also add a domain-oriented prefix sequence of key-value pairs to augment the classical attention layer, which is independently applied to all Transformer layers of pre-trained models to elicit the knowledge interactively and achieve overall optimization. In this case, different domain words from $D_{s}$ and $D_{t}$ can induce relevant domain knowledge while adapting to a new domain. For generalization, we construct discrete prompts using dialogue states or queries, as shown in Figure \ref{fig:intro}, to guide the model to focus on key contents in dialogues and enhance generalization capability on unseen domains. Considering there is no unified and practical benchmark for domain adaptation of dialogue summarization \footnote{Existing work mostly takes news datasets like CNN/DailyMail \cite{Hermann2015TeachingMT} as source domains and dialogue datasets like SAMSum \cite{Gliwa2019SAMSumCA} as target domains. We argue the setting doesn't fit in with industrial scenarios transferring from one business to another, like from attraction consultation to hotel reservation.}, we build domain adaptation benchmarks based on two existing multi-domain summarization datasets TODSum \cite{zhao2021todsum} and QMSum \cite{zhong-etal-2021-qmsum}. Extensive experiments demonstrate the benefits of our methods both in zero-shot and few-shot settings for domain adaptation.



Our contributions are threefold: (1) To the best of our knowledge, we are the first to explore fine-tuning methods for domain adaptation of dialogue summarization task, and establish two practical and comprehensive benchmarks for TODSum and QMSum datasets. (2) We propose a lightweight and efficient Domain-Oriented Prefix-tuning model, with domain word initialized prefix and discrete prompts, to elicit knowledge from large-scale pre-trained models interactively. (3)  We conduct sufficient experiments and qualitative analysis to prove the effectiveness of our methods and discuss current challenges of domain adaptation for dialogue summarization.

\section{Related Work}

\begin{figure*}[t]
\centering
\includegraphics[width=15cm, height=5cm]{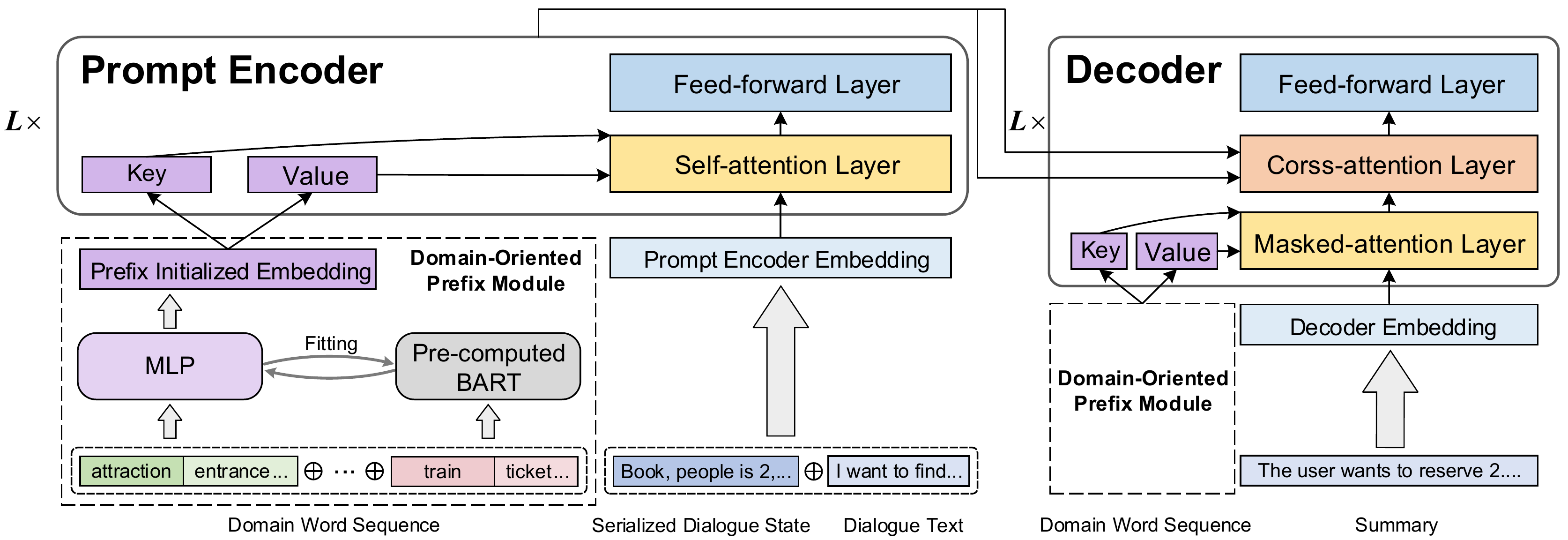}
\caption{Overview of the Domain-Oriented Prefix-tuning model. The input sequence of domain-oriented prefix module includes domain words from four source domains, i.e., \emph{attraction, hotel, taxi, train} and the domain words of \emph{restaurant} domain are used as the prefix sequence of the target domain during the test. The input sequence of prompt encoder is the dialogue state from TODSum dataset and its original dialogue text. The input sequence of decoder is the golden summary. Here, we use the pre-computed outputs of BART as the target of the fitting process to initialize the parameters of MLP. Note that the grey pre-computed BART module does not participate in the training of the model.}
\label{fig:model}
\vspace{-0.5cm}
\end{figure*} 

\noindent\textbf{Abstractive Dialogue Summarization} \quad Dialogue summarization has drawn much attention recently.  For chit-chat scenarios, researchers improved the performance on SAMSum dataset \cite{Gliwa2019SAMSumCA} via topic word information \cite{zhao-etal-2020-improving,liu-etal-2021-topic-aware}, conversational structures \cite{chen-yang-2020-multi,chen2021structureaware}, personal named entity planning \cite{liu-chen-2021-controllable}, and semantic slots \cite{zhao-etal-2021-give-truth}. \citet{Liu2019AutomaticDS}, \citet{Zou2021UnsupervisedSF,Zou2021TopicOrientedSD}, and \citet{lin2021csds} proposed customer-service dialogue summarization datasets under diverse business scenarios. Besides, meeting transcripts, such as AMI \cite{Carletta2005TheAM}, ICSI \cite{1198793}, MediaSum \cite{zhu-etal-2021-mediasum}, and QMSum \cite{zhong-etal-2021-qmsum}, were also studied to promote dialogue summarization technologies. \citet{zhao2021todsum} further proposed a task-oriented dialogue summarization dataset ,TODSum, with dialogue state knowledge. Although great progress has been made in dialogue summarization, few people pay attention to the issue of domain adaptation in dialogue summarization. In this paper, we explore this issue in two multi-domain dialogue summarization datasets, i,e, TODSum and QMSum.

\noindent\textbf{Domain Adaptation in Summarization}\quad \citet{hua-wang-2017-pilot} and \citet{wang2019exploring} adopted the document categories in news publications to build a multi-domain summarization dataset and investigated the domain shift for extractive summarization task. \citet{yang-etal-2020-ted}, \citet{Zhang_2020_ICML}, \citet{magooda2020abstractive}, and \citet{fabbri-etal-2021-improving} regarded diverse summarization datasets as different domains and conducted the assessment of multi-domain settings. Furthermore, various stages of pre-trainings were added to narrow the gap between the pre-training in news domain and the fine-tuning in dialogue domain \cite{yu-etal-2021-adaptsum,zou-etal-2021-low}. However,  these methods focus on the heavy pre-training stage rather than the lightweight fine-tuning, which is time-consuming and relies on large-scale external corpora. Therefore, we try to explore the fine-tuning methods for domain adaptation of dialogue summarization task.

\noindent\textbf{Prompts in Summarization} \quad With the arrival of GPT-3, prompt learning has become a nascent field, which conducts task-specific adaptation of large language models (LMs) via prepending an instruction. \citet{schick2021fewshot} explored the fixed-prompt LM tuning for few-shot text summarization with manually crafted templates. \citet{zhao2021todsum} and \citet{dou2020gsum} further adopted the prompt+LM tuning strategy on text summarization task, where learnable prefix prompts are different types of guidance signals. \citet{li2021prefix} investigated fixed-LM prompt tuning, where learnable prefix tokens are prepended to the input while parameters in pre-trained models are frozen. Following \citet{li2021prefix}, we design the domain information to initialize the continuous prefix module, and use discrete prompts and dialogue texts to optimize prefix parameters, which greatly reduces the size of parameters and is suitable for low-resource scenarios.

\section{Problem Formulation}


Domain adaptation of dialogue summarization aims to generate the summary $\mathbf{y}$ conditioned on the source dialogue $\mathbf{x}_{d}$, where the training and test are in different domains. We add the domain word prefix $\mathbf{x}_{dw}$ and discrete prompt $\mathbf{x}_{dp}$ as additional input which we will describe details later. Note that we only update the prefix-related parameters and fix the parameters of BART. We train the model using the source data and test using the target data.

\section{Methodology}

As Figure \ref{fig:model} shows,  our model is on the basis of the framework of BART, including a domain-oriented prefix module, a prompt encoder, and a decoder.

\subsection{Domain-oriented Prefix}
\vspace{-0.1cm}


To alleviate domain entanglement, we present a domain-oriented prefix module to obtain the shared knowledge of the source domain $D_s$ and the target domain $D_t$. It is designed as follows:

\noindent\textbf{Initialization} \quad We extract some keywords from dialogue texts in each domain by LDA \cite{NIPS2010_3902} and concatenate them all together as a domain word (prefix) sequence $\mathbf{x}_{dw}$\footnote{We represent some of domain words for each domain in Appendix \ref{sec:domain_words}.}. Randomly initialized embeddings of the domain word sequence compose a learnable matrix $M_{\theta}\in\mathbb{R}^{|\mathbf{x}_{dw}|\times d_m}$.


\noindent\textbf{Parametrization} \quad We use an MLP to encode the domain-oriented prefix module, which stably elicits knowledge from the large pre-trained model in the prefix-tuning process. Specifically, we first input the domain word sequence to the MLP and the pre-computed BART respectively, then re-train the MLP by fitting its outputs with the decoder hidden states of the pre-computed BART using MSE loss. In this fitting process, we only iteratively update MLP parameters $\varphi \in \mathcal{R}^{d_m \times d_n}$ and keep the pre-computed BART fixed. Finally, we get the initialization parameters of MLP and use this pre-trained MLP to map the initialized embeddings of prefix representations for each Transformer layer both in prompt encoder and decoder:

\vspace{-0.1cm}
\begin{equation}
\begin{aligned}
M_{\theta}^{'}[i,:]={\operatorname{MLP}_{\varphi}}(M_{\theta}[i,:])
\end{aligned}
\end{equation}
where $i\in \mathbf{x}_{dw}$ and $M_{\theta}^{'}\in\mathbb{R}^{|\mathbf{x}_{dw}|\times d_n}$. 
Note that this continuous prefix is applied for every layer of the large-scale pre-trained model independently.

\subsection{Prompt Encoder}

\noindent\textbf{Discrete Prompts}  \quad We utilize some discrete prompts to emphasize key elements in dialogues and enhance the model generalization to new domains. Here, discrete prompts are dialogue states of TODSum dataset or queries of QMSum. Considering that the original form of dialogue states is \emph{book (people=5; day=Monday)} which is not compatible with BART encoder, we convert this structured information into a serialized sequence, i.e., \emph{book, people is 5, day is Monday}, to improve the stability of training. Note that we do not make any changes to the query of QMSum dataset because it is already a serialized representation.

For prompt encoder, we firstly concatenate the discrete prompt sequence $\mathbf{x}_{dp}$ and dialogue text sequence $\mathbf{x}_d$ as the input sequence of encoder, i.e., $\mathbf{x}_{enc}=[\mathbf{x}_{dp};\mathbf{x}_{d}]$. Then, the $\mathbf{x}_{enc}$ is fed into the prompt encoder based on the BART encoder, containing multiple Transformer layers. Note that we modify the self-attention mechanism by adding a domain-oriented prefix sequence of key-value pairs, which learns the knowledge from the pre-trained model through interactions with the dialogue text to carry out the overall task. For the typical $l_e$-th Transformer layer in encoder, the query ($\mathbf{Q}_{l_e}$), key ($\mathbf{K}_{l_e}$), and value ($\mathbf{V}_{l_e}$) matrices are computed through linear transformations on the hidden states of $\mathbf{x}_{enc}$. Here, we further augment the $\mathbf{K}_{l_e}$ and $\mathbf{V}_{l_e}$:
\begin{equation}
\begin{aligned}
\mathbf{K}_{l_e}^{'}=[\mathbf{P}_{l_e,k};\mathbf{K}_{l_e}], \  \mathbf{V}_{l_e}^{'}=[\mathbf{P}_{l_e,v};\mathbf{V}_{l_e}]
\end{aligned}
\end{equation}
where $\mathbf{P}_{l_e,k}$, $\mathbf{P}_{l_e,v}$ are computed through linear transformations on $M_{\theta}^{'}$. $\mathbf{K}_{l_e}^{'}, \mathbf{V}_{l_e}^{'}\in \mathbb{R}^{(|\mathbf{x}_{dw}|+|\mathbf{x}_{dp}|+|\mathbf{x}_{d}|)\times d_n}$ and ${l_e}$$\in$$L$. The augmented self-attention layer is finally calculated as follows:
\begin{equation}
\begin{aligned}
\mathbf{A}_{self}=\operatorname{softmax}(\mathbf{Q}_{l_e}{\mathbf{K}_{l_e}^{'}}^{T})\mathbf{V}_{l_e}^{'}
\end{aligned}
\end{equation}


\subsection{Decoder}

We also prepend the prefix module for decoder, where the cross-attention and masked-attention mechanisms are augmented in a similar way. The cross-attention between the $l_e$-th layer of prompt encoder and the $l_d$-th layer of the decoder is designed as:
\begin{equation}
\begin{aligned}
\mathbf{A}_{cross}=\operatorname{softmax}(\mathbf{Q}_{l_d}{\mathbf{K}_{l_e}^{'}}^{T})\mathbf{V}_{l_e}^{'}
\end{aligned}
\end{equation}
where $\mathbf{Q}_{l_d}$ is computed through a linear transformation on the hidden states of the summary text $\mathbf{x}_{s}$ and ${l_d}$$\in$$L$. Besides, the implementation of masked-attention layer is the same as the self-attention layer in the prompt encoder.



\subsection{Training Strategy}
In the domain-oriented prefix module, the parameter set of all linear transformations is symbolized as $\alpha$. For training strategy in DOP, we perform gradient updates on the following log-likelihood objective:
\begin{equation}
\begin{aligned}
\underset{\alpha,\theta,\varphi}{\mathop{\max \ }} {\log}\,p_{\alpha,\theta,\varphi, \phi}(\mathbf{y}|\mathbf{x})=\sum_{i\in
\mathbf{y}}{\log}\,p_{\alpha,\theta,\varphi,\phi}(y_i|y_{<i})
\end{aligned}
\end{equation}
where the BART parameters $\phi$ are fixed. The prefix parameters $\alpha$, $\theta$, and $\varphi$ are the only trainable parameters. During the training, we use the domain words from source domains as the prefix sequence. When the training is completed, we save all parameters of the domain-oriented prefix module and drop the pre-computed BART module. During the test, the domain words from the target domain are mapped to the representations of prefixes only via the reserved $\operatorname{MLP}_{\varphi}(\cdot)$, where the features of the source domain $D_s$ can be transferred to the target domain $D_t$.



\section{Experimental Setup}

\subsection{Datasets}

\begin{table}[]
\centering
\resizebox{0.48\textwidth}{!}{%
\begin{tabular}{l|c|c|c|c}
\hline
\textbf{Domains}  & \textbf{Size}      & \textbf{Dialog.len}               & \textbf{Summ.len}      & \textbf{DS.len}  \\ 
\hline
\textbf{Train} & 345        & 120.67      & 24.93       & 18.29 \\
\textbf{Taxi} & 435         & 80.24      & 29.04        & 15.80 \\
\textbf{Restaurant} & 1,311  & 105.42      & 23.04       & 14.30  \\
\textbf{Hotel}  & 636       & 145.16      & 30.06       & 21.38 \\
\textbf{Attraction}  & 150  & 95.48      & 22.27        & 7.92 \\
\textbf{All}    & 2,877      & 111.71    & 25.68         & 16.24\\
\hline
\end{tabular}
}
\vspace{-0.25cm}
\caption{Details of TODSum. "Dialog.len" denotes the average length of dialogues, "Summ.len" denotes the average length of summaries, and "DS.len" denotes the average length of serialized dialogue states.}
\label{tab:dataset-todsum}
\end{table}

\begin{table}[]
\centering
\resizebox{0.48\textwidth}{!}{%
\begin{tabular}{l|c|c|c|c}
\hline
\textbf{Domains}  & \textbf{Size}      & \textbf{Dialog.len}               & \textbf{Summ.len}      & \textbf{QR.len}  \\ 
\hline
\textbf{Academic} & 312        & 1,155.78     & 46.48       & 8.56 \\
\textbf{Committee} & 417         & 757.68      & 76.00        & 14.54 \\
\textbf{Product} & 847       & 971.65       & 63.96      & 13.36  \\
\textbf{All}    & 1,576      & 951.49      & 63.68         & 12.73\\
\hline
\end{tabular}
}
\vspace{-0.25cm}
\caption{Details of QMSum. "QR.len" denotes the average length of queries.}
\label{tab:dataset-qmsum}
\vspace{-0.5cm}
\end{table}

\begin{table*}[]
\centering
\resizebox{1.0\textwidth}{!}{%
\begin{tabular}{l|ccc|ccc|ccc|ccc|ccc}
\hline
\multirow{3}{*}{\textbf{Models}} & \multicolumn{3}{c|}{\textbf{Train}} & \multicolumn{3}{c|}{\textbf{Taxi}} & \multicolumn{3}{c|}{\textbf{Restaurant}} & \multicolumn{3}{c|}{\textbf{Hotel}} & \multicolumn{3}{c}{\textbf{Attraction}} \\ \cline{2-16} &
\multicolumn{3}{c|}{\textbf{2,332 / 200 / 345}}&
\multicolumn{3}{c|}{\textbf{2,242 / 200 / 435}}&
\multicolumn{3}{c|}{\textbf{1,366 / 200 / 1,311}}&
\multicolumn{3}{c|}{\textbf{2,041 / 200 / 636}} &
\multicolumn{3}{c}{\textbf{2,527 / 200 / 150}} \\ \cline{2-16} 
                                 & \textbf{R-1}  & \textbf{R-2} & \textbf{R-L} & \textbf{R-1} & \textbf{R-2} & \textbf{R-L} & \textbf{R-1}   & \textbf{R-2}   & \textbf{R-L}   & \textbf{R-1}  & \textbf{R-2} & \textbf{R-L} & \textbf{R-1}   & \textbf{R-2}   & \textbf{R-L}  \\ \hline\hline
\textbf{Lead-3} & 20.36 & 2.78 & 16.07 & 24.20 & 7.34 & 20.75 & 28.27 & 6.10 & 23.49 & 23.86 & 4.58 & 18.80 & 22.76 & 5.28 & 19.66 \\
\textbf{Oracle} & 39.06 & 10.04 & 32.87 & 38.96 & 14.06 & 33.43 & 45.79 & 15.57 & 38.42 & 39.65 & 11.28 & 32.56 & 41.90 & 14.18 & 38.79 \\
\textbf{BertExt}                 & 39.19         & 9.71         & 33.24        & 38.49        & 13.57        & 33.36        & 40.64          & 12.34          & 34.43          & 35.96         & 9.71         & 30.10        & 36.25          & 11.19          & 31.41         \\ \hline
\textbf{PGN}                     &  32.50             & 10.47             & 29.33             & 32.48             & 7.79             &  29.82            & 33.63               & 10.78               & 31.47                & 32.18              & 9.36             & 30.93              & 32.66               & 9.95               & 30.29              \\
\textbf{Transformer}             &  33.47             &      10.98        &    30.28          &     33.35         &         8.71     &    30.57          &      34.49         &      11.43          &   31.99             &     33.05          &    10.62          &     31.63         &   33.18             & 10.74                &   30.91            \\
\textbf{BertAbs}                 & 42.89         & 16.57        & 37.32        & 36.43        & 14.69        & 32.15        & 42.10           & 18.61          & 38.87          & 38.03         & 13.34        & 33.22        & 36.21          & 14.81          & 34.67         \\
\textbf{BART}                    & 46.82         & 18.42        & 42.06        & 39.98        & 15.79        & 34.41        & 47.02          & 22.62          & 44.93          & 40.84         & 14.20        & 36.83        & 43.67          & 20.23          & 41.44         \\
\textbf{BART w. DS}              & 49.02         & 23.80        & 44.59        & 43.59        & 19.56        & 38.65        & 49.25          & 23.57          & 45.23          & 43.97         & 17.02        & 39.31        & 47.55          & 22.62          & 45.16         \\
\textbf{Prefix-tuning}           & 45.92         & 22.70        & 41.06        & 41.89        & 19.47       & 39.62        & 47.19          & 24.20           & 42.99          & 43.41         & 18.75        & 36.75        & 44.48          & 22.43          & 40.94         \\
\textbf{DOP (ours)}              & \textbf{52.51}         & \textbf{25.45}        & \textbf{47.78}       & \textbf{47.14}        & \textbf{24.37}        & \textbf{42.75}        & \textbf{51.28}          & \textbf{32.68}          & \textbf{47.44}         & \textbf{48.44}         & \textbf{24.58}        & \textbf{41.45}        & \textbf{52.90}          & \textbf{30.51}          & \textbf{49.48}         \\ \hline
\end{tabular}%
}
\caption{ROUGE scores of the zero-shot setting for TODSum. Results are averaged over three random runs. "DS" denotes the dialogue states. Values in the second row denote the size of train/valid/test set. ($p < 0.05$ under t-test)}
\label{main_results_todsum}
\vspace{-0.3cm}
\end{table*}

\begin{table}[]
\centering
\resizebox{0.5\textwidth}{!}{%
\begin{tabular}{l|ccc|ccc|ccc}
\hline
\multirow{3}{*}{\textbf{Models}} & \multicolumn{3}{c|}{\textbf{Academic}} & \multicolumn{3}{c|}{\textbf{Committee}} & \multicolumn{3}{c}{\textbf{Product}} \\ \cline{2-10} &
\multicolumn{3}{c|}{\textbf{1,069 / 195 / 312}} &
\multicolumn{3}{c|}{\textbf{981 / 178 / 417}} & 
\multicolumn{3}{c}{\textbf{614 / 115 / 847}}
\\ \cline{2-10} 
                                 & \textbf{R-1}   & \textbf{R-2}  & \textbf{R-L}  & \textbf{R-1}   & \textbf{R-2}   & \textbf{R-L}  & \textbf{R-1}  & \textbf{R-2}  & \textbf{R-L} \\ \hline\hline
\textbf{Lead-3} & 14.86 & 2.68 & 12.62 & 23.18 & 7.71 & 19.33 & 18.61 & 4.20 & 15.55\\
\textbf{Oracle} & 32.44 & 8.78 & 27.92 & 54.92 & 35.03 & 51.85 & 40.23 & 15.29 & 35.83 \\
\textbf{BertExt}                 & 19.36          & 2.72          & 16.90         & 31.08          & 11.32          & 27.46         & 23.43         & 4.93          & 20.63        \\ \hline
\textbf{PGN}    & 19.38           &  2.65             &  17.06              &  20.14              &   3.30            &     18.22          &      17.59         &    2.07 & 15.21          \\
\textbf{Transformer} & 18.72               &  2.53             &   16.97            &     20.01           & 3.16                &        17.88       &   17.14            &         2.02      & 15.14              \\
\textbf{BertAbs}                 & 20.36          & 2.45          & 17.84         &   20.93             &   3.23             &     18.26          &  17.32             &   1.86            &  15.30            \\
\textbf{BART}                    & 20.53          & 3.63          & 18.09         & 26.65          & 8.28           & 24.84         & 23.14         & 5.12          & 20.87        \\ 
\textbf{BART w. QR}              & 22.45          & 4.11          & 20.14         & 27.93          & 9.73           & 25.87         & 24.89         & 5.93          & 22.39        \\
\textbf{Prefix-tuning}           & 21.90               & 4.01              &  18.71             &  27.70              &    9.37            & 25.62              &  25.77             &    7.26           &   23.06           \\
\textbf{DOP}              & \textbf{24.43}          & \textbf{5.84}          & \textbf{21.49}         & \textbf{30.28}          & \textbf{11.63}          & \textbf{27.33}         & \textbf{29.85}         & \textbf{9.41}          & \textbf{26.88}        \\ \hline
\end{tabular}%
}
\vspace{-0.1cm}
\caption{ROUGE scores of the zero-shot setting for QMSum. "QR" denotes the queries. Same as TODSum, values in the second row denote the size of train/valid/test set. All results are averaged over three random runs. ($p < 0.05$ under t-test)}
\label{main_results_qmsum}
\vspace{-0.6cm}
\end{table}

We evaluate our model on two multi-domain dialogue summarization datasets and the details of statistics are shown in Table \ref{tab:dataset-todsum} and Table \ref{tab:dataset-qmsum}:

\noindent\textbf{TODSum}: This dataset is proposed by \citet{zhao2021todsum}, which is a task-oriented dialogue summarization dataset based on the classic dialogue dataset MultiWOZ \cite{budzianowski2018multiwoz}. According to the domain information, the dataset can be divided into five domains: \emph{restaurant}, \emph{hotel}, \emph{attraction}, \emph{taxi}, and \emph{train}. Considering that parts of dialogues in TODSum contain multiple domains, in this paper, we firstly select all single-domain dialogues from TODSum as the dataset used for this experiment. Then we integrate all these samples in any four of five domains as source domains $D_s$ and the other one is regarded as target domain $D_t$, where 200 samples extracted from $D_s$ are as the validation set, the remaining as the training set, and samples of $D_t$ are the test set.


\noindent\textbf{QMSum}: This dataset is proposed by \citet{zhong-etal-2021-qmsum}, which contains hundreds of meeting transcriptions and includes three domains, i.e., \emph{academic}, \emph{committee}, \emph{product}. In addition, each sample can be divided into several dialogue fragments according to some queries and the answer for the corresponding query is its golden summary. Such \emph{(query-dialogue-answer)} pairs are usually used for query-based meeting summarization tasks. In this paper, we separately integrate the training and validation sets, and test set in any two of three domains as the training data and validation data, i.e., the data of source domain $D_s$. All data of the other domain is used as the test data, i.e., the data of target domain $D_t$.

\subsection{Baselines and Evaluation Metrics}

We compare our methods with saveral baselines. The extractive baselines are included: (1) Lead-3; (2) Oracle; (3) BertExt \cite{liu2019text}. Some abstractive methods are also added for comparison: (1) PGN \cite{see-etal-2017-get}; (2) Transformer \cite{10.5555/3295222.3295349}; (3) BertAbs \cite{liu2019text}; (4) BART \cite{Lewis2020BARTDS}; (5) BART w. DS/QR \cite{zhao2021todsum}; (6) Prefix-tuning \cite{li2021prefix}. For QMSum, we also introduce its benchmark \cite{zhong-etal-2021-qmsum}. Since this method feeds the extracted spans into BART, we integrate the results of this method with the results of BART. We use ROUGEs \cite{lin2004rouge,lin2004automatic} to quantitatively evaluate the performance of our methods. Our codes are publicly available \footnote{https://github.com/Zeng-WH/DOP-Tuning.}.
We give the baselines and evaluation metrics in Appendix \ref{baselines} and Appendix \ref{metric}.

\subsection{Training Details}
Our implementation is based on the Hugging Face Transformer models\footnote{https://github.com/huggingface/transformers}. $\rm BART_{LARGE}$ is used as a backbone and the source dialogue sequence is truncated to 1024 BPE tokens. For domain-oriented prefix module, the MLP maps 1024 dimension into 24576 dimension, which is calculated by $2*\operatorname{number \ of\ decoder\ layers}*1024$ and the numbers of domain words (prefix length) are set to 140 and 200 for TODSUM and QMSum datasets. Following the settings in \citet{li2021prefix}, we use an AdamW optimizer \cite{loshchilov2018decoupled} and a linear learning rate scheduler with initial rate of $5\cdot10^{-5}$, and the batch size is set to 5. Our model is trained on RTX 2080 Ti machines, taking only 5 minutes per epoch on TODSum dataset and 3 min per epoch on QSum dataset. However, BART w. DS takes 13 minutes and 8 minutes per epoch on TODSum and QMSum datasets. The reason for the shorter training time of our model is that the trainable number of parameters of the DOP model is only 15\% of the BART w. DS. For all experiments, we set the number of training epochs to 30. At the decoding phase, we use a beam size of 6 and max generated tokens of 125. The decoding takes 1.3 seconds per batch with the size of 2.



\subsection{Main Results}
\noindent\textbf{Results on TODSum} \quad
Table \ref{main_results_todsum} presents the results of the zero-shot setting for TODSum dataset, where each of the five domains is regarded as the target domain respectively. The division of dataset in the second row intuitively shows that the amount of data in $D_s$ is small and limited. We conduct experiments based on some common extractive models and some strong abstractive baselines. We also add a lightweight fine-tuning summarizer for comparison. As observed, for most ROUGEs, Prefix-tuning performs worse than BART and BART w. DS. It is because the dialogue text is long and complex, and using only 20\% parameters of fine-tuning can not well understand the domain knowledge and identify the key contents in dialogues. Compared to Prefix-tuning of the same magnitude parameters, our model improves by 7\%, 3\%, 7\% for \emph{train} domain, 5\%, 5\%, 3\% for \emph{taxi} domain, 4\%, 8\%, 4\% for \emph{restaurant} domain, 5\%, 6\%, 5\% for \emph{hotel} domain, and 8\%, 8\%, 9\% for \emph{attraction} domain. This shows that the prefix module initialized by domain words and the discrete prompts composed of dialogue states play important roles. Besides, our model still surpasses BART w. DS, a full-parameter fine-tuning based model, which further illustrates that our model efficiently disentangles the knowledge of the source and target domains. Note that \emph{attraction} domain gets the highest ROUGEs and the increased margins are also the largest. This may be due to the high overlaps between the \emph{attraction} and source domains. All the results suggest that with limited data, the performance of our model still reaches state-of-the-art.

\noindent\textbf{Results on QMSum} \quad Table \ref{main_results_qmsum} displays the results on zero-shot out-domain tests in three domains of QMSum dataset. As seen from the second row, in addition to the limited data, the source domain size may be even less than the target domain size, i.e., \emph{product} domain. The trend of overall performance is consistent with that of TODSum dataset \footnote{Note that ROUGEs of Oracle are very high in QMSum, which is because most parts of the golden summaries are directly copied from the original dialogue. It is determined by the characteristics of the QMSum and the results are consistent with \citet{zhong-etal-2021-qmsum}.}, where the improvement in \emph{product} domain is the most obvious and there are 5\%, 4\% and 5\% increased for R-1, R-2, and R-L, respectively. However, all the ROUGEs are low, which is because there are no obvious domain words, leading to serious domain entanglement. Besides, due to the longer meeting text, it is hard to capture the key contents in dialogues, so as to the poor generalization in the target domain. Generally, these results show that the multi-domain setting in meeting summarization task is apparently necessary and meaningful. Meeting transcripts cover various domains, making the adaptation of models particularly difficult.

\section{Qualitative Analysis}

\subsection{Effect of Domain Words}

\begin{figure}[t]
    \centering
    \subfigure[Number of domain words]{
        \includegraphics[scale=0.215]{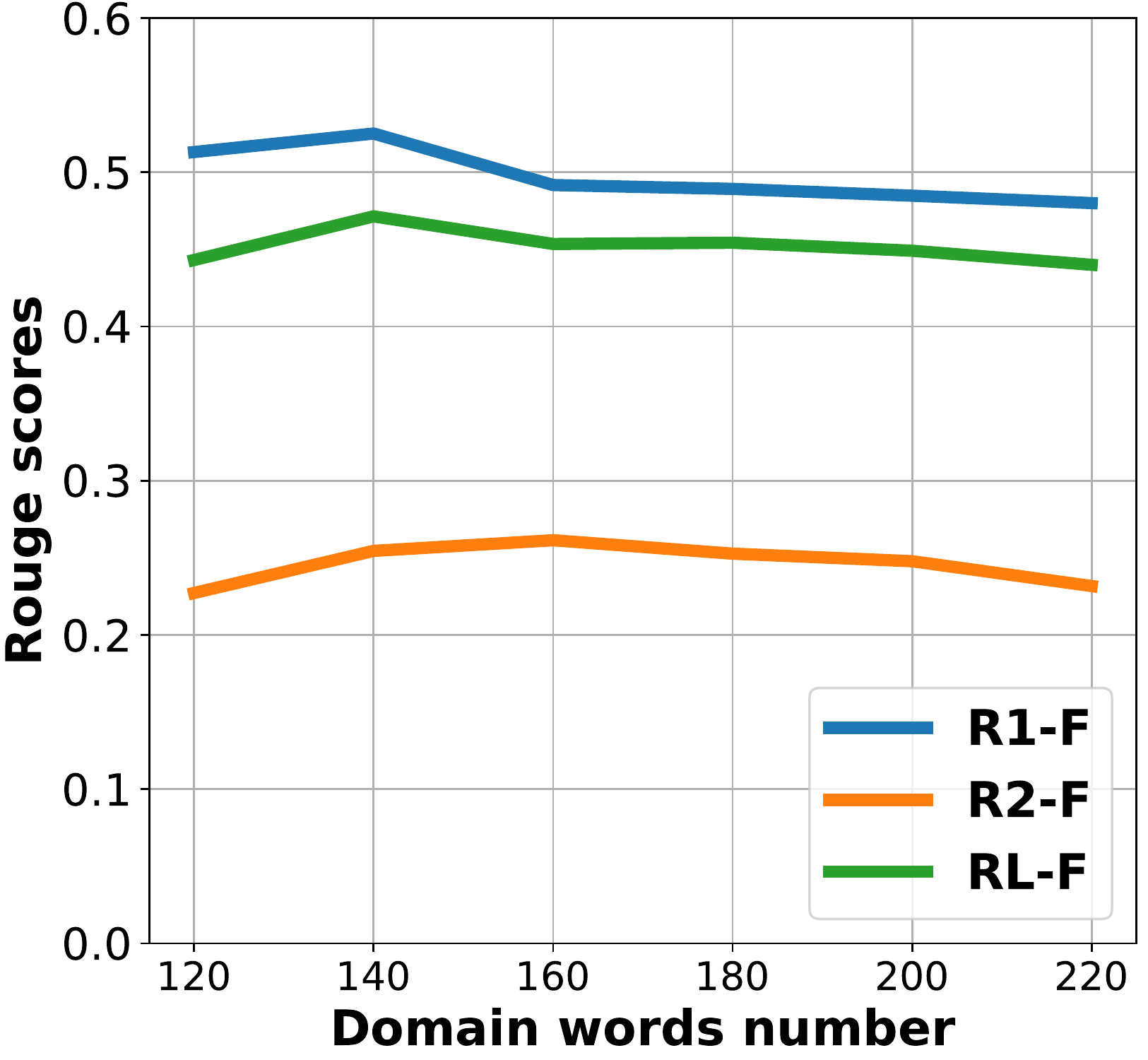}
    }
    \subfigure[Quality of domain words]{
        \includegraphics[scale=0.215]{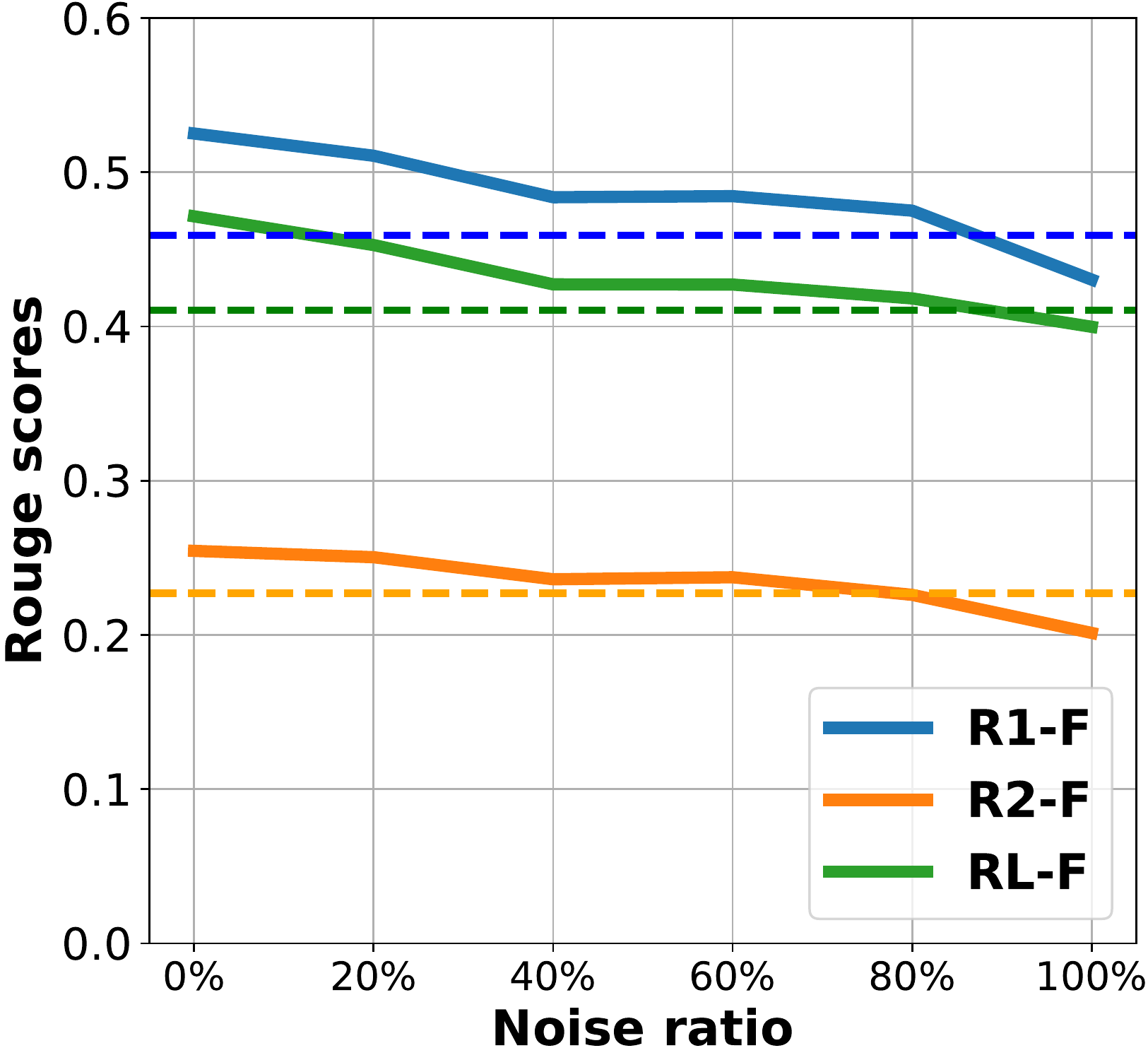}
    }
    \vspace{-0.3cm}
    \caption{Effect of different numbers and different qualities of domain words for \emph{train} domain in TODSum. The dotted lines in (b) respectively represent the F1 sources of  ROUGE-1, ROUGE-2, and ROUGE-L of the Prefix-tuning model.}
    \label{prefix}
    \vspace{-0.5cm}
\end{figure}

\noindent\textbf{Number of Domain Words} \quad We set different numbers of domain words, i.e., prefix length, to test the performance of our model DOP for \emph{train} domain in TODSum dataset. As shown in Figure \ref{prefix} (a), among these setting candidates, there is a threshold (140) that allows the ROUGEs to reach the peak. When choosing a fewer setting, the model does not perform well due to insufficient number of parameters, which is improved as the number increases. When being more than this threshold, a drop in performance occurs. One reason is that too long a sequence adds a large burden to BART, and the other one is that it introduces excessive noise. However, the change in the number of the domain words does not have a great impact on the performance (only 2$\sim$3\% fluctuation), which also reflects the effectiveness of domain-oriented prefix module and the robustness of our model.

\noindent\textbf{Quality of Domain Words} \quad For \emph{train} domain in TODSum, we randomly replace a certain percentage of the domain words with words that are not related to the source domain. As Figure \ref{prefix} (b) shows, when more noise is introduced, the model suffers more interference and its performance decreases. However, it performs better than Prefix-tuning. Only when the proportion of noise reaches 100\%, the performance of our model is even worse than that of Prefix-tuning. This is because we especially use completely irrelevant words for initialization and fitting, which introduces more noise than simple random initialization and affects the performance of DOP. From this point of view, introducing high-quality domain words is good for domain disentanglement and the quality of domain words is critical to summary generation.

\subsection{Ablation Study}
We perform ablation experiments on \emph{train} domain of TODSum dataset and \emph{committee} domain of QMSum dataset, as shown in Tables \ref{tab:ablation_DOTSum} and \ref{tab:ablation_QMSum}. We can observe that the removal of domain-oriented initialization in the prefix module will make the ROUGEs decrease significantly. Especially for TODSum, R-1, R-2, and R-L drop by 4\%, 2\%, and 3\%, which shows the importance of domain word information for inducing the relevant knowledge while adapting to a new domain. In addition, after we remove the discrete prompts,  i.e. dialogue state and query, the performance of the models becomes worse, but still outperforms the results of Prefix-tuning. It demonstrates that discrete prompts help the model pay attention to the key elements in the dialogue and improve the generalization of the model. Notably, our model achieves summary generation only by optimizing the domain-oriented prefix module, where domain words are available in all datasets. Since the DS and QR features happen to exist in the two datasets, we take advantage of them together with dialogue texts. When removing both DW and DS/QR at the same time, the model is equivalent to Prefix-tuning and the results are consistent.

\begin{table}[t]
\centering
\resizebox{0.43\textwidth}{!}{%
\begin{tabular}{l|ccc}
\hline
\textbf{Model}               & \textbf{R-1}   & \textbf{R-2}   & \textbf{R-L}   \\ \hline
\textbf{DOP (ours)}               & 52.51 & 25.45 & 47.78 \\
\quad \textbf{w/o DW}   & 48.87 & 23.81 & 44.52\\ 
\quad \textbf{w/o DS}   & 47.59 & 23.25 & 43.41\\
\quad  \textbf{w/o DW \& DS} & 45.92 & 22.70 & 41.06\\ \hline
\end{tabular}%
}
\vspace{-0.1cm}
\caption{F1 scores of ablation study on \emph{train} domain of TODSum dataset. "DW" denotes domain words and "DS" denotes dialogue states.}
\label{tab:ablation_DOTSum}
\vspace{-0.1cm}
\end{table}

\begin{table}[t]
\centering
\resizebox{0.43\textwidth}{!}{%
\begin{tabular}{l|ccc}
\hline
\textbf{Model}               & \textbf{R-1}   & \textbf{R-2}   & \textbf{R-L}   \\ \hline
\textbf{DOP (ours)}               & 30.28 & 11.63 & 27.33 \\
\quad \textbf{w/o DW}   & 29.10 & 10.59 & 26.53\\ 
\quad \textbf{w/o QR}   & 28.47 & 9.60 & 26.38\\
\quad  \textbf{w/o DW \& QR} & 27.70 & 9.37 & 25.62\\ \hline
\end{tabular}%
}
\caption{F1 scores of ablation study on \emph{committee} domain of QMSum dataset. "DW" denotes domain words and "QR" denotes queries.}
\label{tab:ablation_QMSum}
\vspace{-0.45cm}
\end{table}

\subsection{Effect of Prefix Module in Encoder and Decoder}
Since both the encoder and the decoder in our DOP introduce the prefix module, we verify their effects in the \emph{train} and \emph{committee} domains respectively. As shown in Table \ref{encdec prefix}, when the encoder prefix module or decoder prefix module is removed, the performance of the model decreases, which shows that both are necessary and effective. In addition, we find that it is interesting that removing the prefix on the encoder side has a smaller impact on the model than removing the decoder side, especially in TODSum (about 5\% on R-1). A reasonable explanation is that the prefix modules in encoder and decoder are responsible for different tasks. The prefix module on the encoder side assists the model to understand the dialogue, while the prefix module on the decoder side assists in model generation. Therefore, for summary generation, the prefix module in decoder is more helpful to the model.

\begin{table}[]
\centering
\resizebox{0.48\textwidth}{!}{
\begin{tabular}{c|l|ccc}
\hline
\textbf{Domain}                     & \textbf{Model} & \textbf{R-1} & \textbf{R-2} & \textbf{R-L} \\ \hline
\multirow{3}{*}{\textbf{Train}}     
& \textbf{DOP}            & 52.51        & 25.45        & 47.14        \\
& \quad \textbf{w/o enc.prefix} & 50.69        & 23.58        & 45.98        \\
& \quad \textbf{w/o dec.prefix} & 45.51        & 22.15        & 40.67        \\ \hline
\multirow{3}{*}{\textbf{Committee}} 
& \textbf{DOP}            & 30.28        & 11.63        & 27.33        \\
& \quad \textbf{w/o enc.prefix} & 29.20        & 11.34        & 26.37        \\
& \quad \textbf{w/o dec.prefix} & 29.11        & 11.29        & 26.30        \\ \hline
\end{tabular}}
\caption{Effects of prefix module in encoder and decoder. We removed the prefix module from the encoder and decoder respectively to verify its effectiveness.}
\label{encdec prefix}
\end{table}

\vspace{-0.2cm}
\subsection{Effect of Training Data}
\vspace{-0.1cm}


\begin{figure}[t]
\centering
\resizebox{0.45\textwidth}{!}{
\includegraphics[scale=0.6]{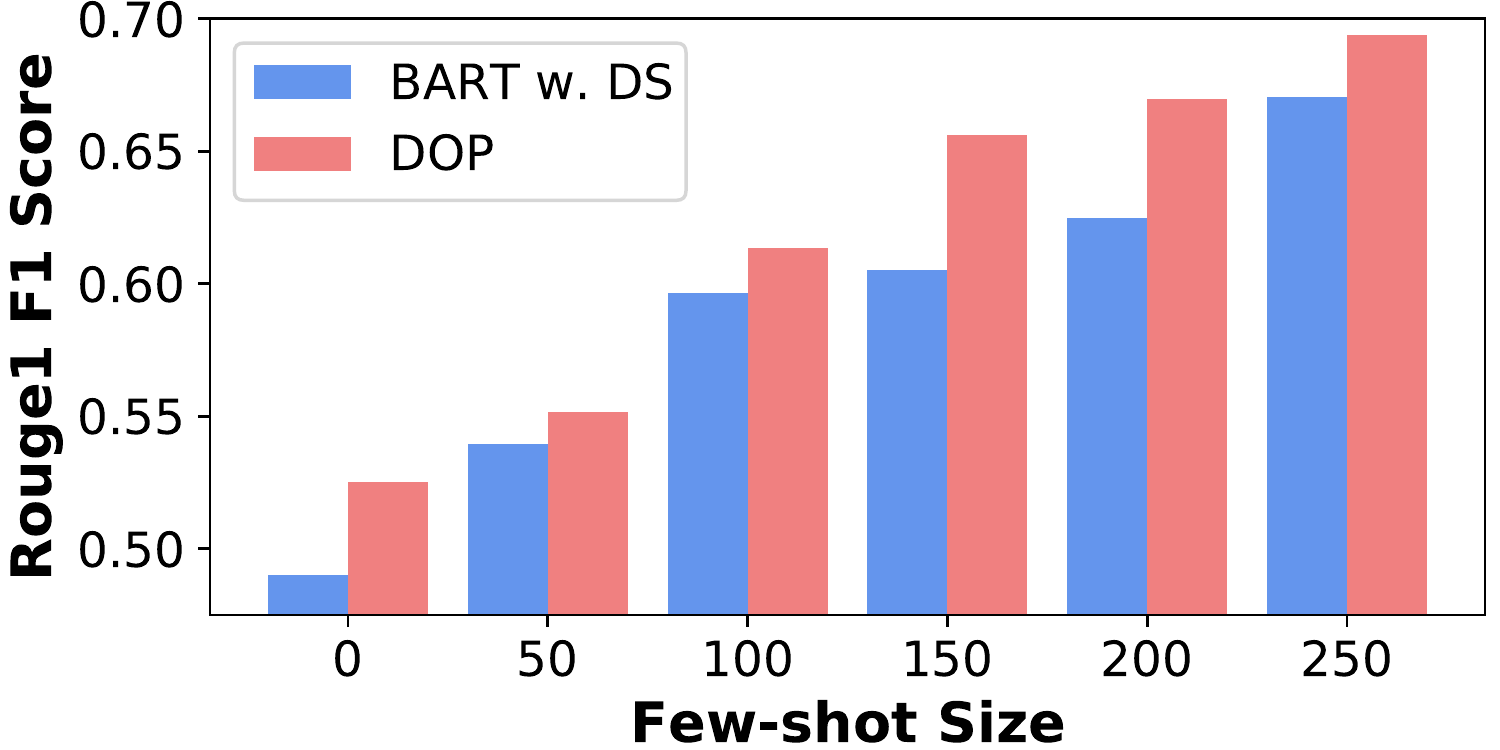}}
\vspace{-0.3cm}
\caption{Effect of different few-shot data size for \emph{train} domain in  TODSum dataset. "0" denotes the zero-shot setting.}
\label{fig:few_shot_size}
\vspace{-0.6cm}
\end{figure}

\begin{figure}[t]
\centering
\resizebox{0.45\textwidth}{!}{
\includegraphics[scale=0.6]{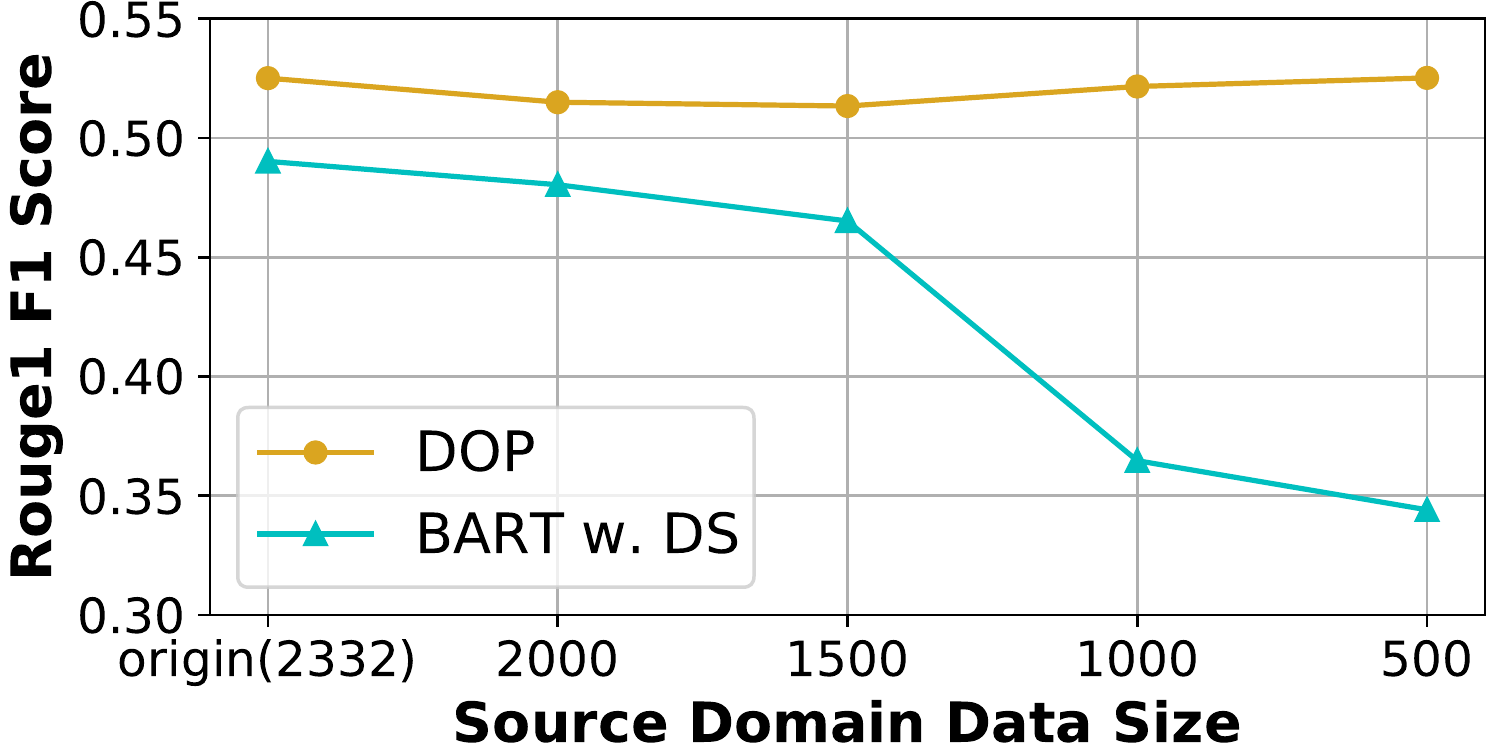}}
\vspace{-0.2cm}
\caption{Effect of different source domain data size for \emph{train} domain in TODSum dataset. "origin (2332)" denotes the same setting as the main results.}
\label{fig:source_domain_size}
\vspace{-0.5cm}
\end{figure}

\noindent\textbf{Performance in Few-shot Settings} \quad For TODSum, we fix the size of source domain data, adding a certain amount of target (\emph{train}) domain data for training, as shown in Figure \ref{fig:few_shot_size}. As the size of target domain data increases, the performance of both BART w. DS and DOP present a steady improvement trend and that of our DOP model is consistently better than BART w. DS, which is as expected. Besides, there is a substantial improvement from 50 to 100. This phenomenon shows that adding target knowledge can help the model learn about information of target domain and after adding a certain amount will help the model more efficiently. 

\noindent\textbf{Effect of Source Domain Data Size} \quad We keep the zero-shot setting unchanged and adjust the size of source domain data for training to observe changes in the performance of the two models for \emph{train} domain in TODSum. As shown in Figure \ref{fig:source_domain_size}, the smaller of data size, the greater the difference between the performance of the DOP and BART w. DS, that is, the performance of BART w. DS is getting worse, while the DOP maintains excellent performance steadily. This demonstrates that our DOP model is insensitive to the data scale and robustness to a certain extent. This also confirms that in the main experiment, our model can be outstanding in very limited and uneven data.

\subsection{Prefix Length vs. Input Length}

\begin{figure}[t]
    \centering
    \subfigure[TODSum]{
        \includegraphics[scale=0.25]{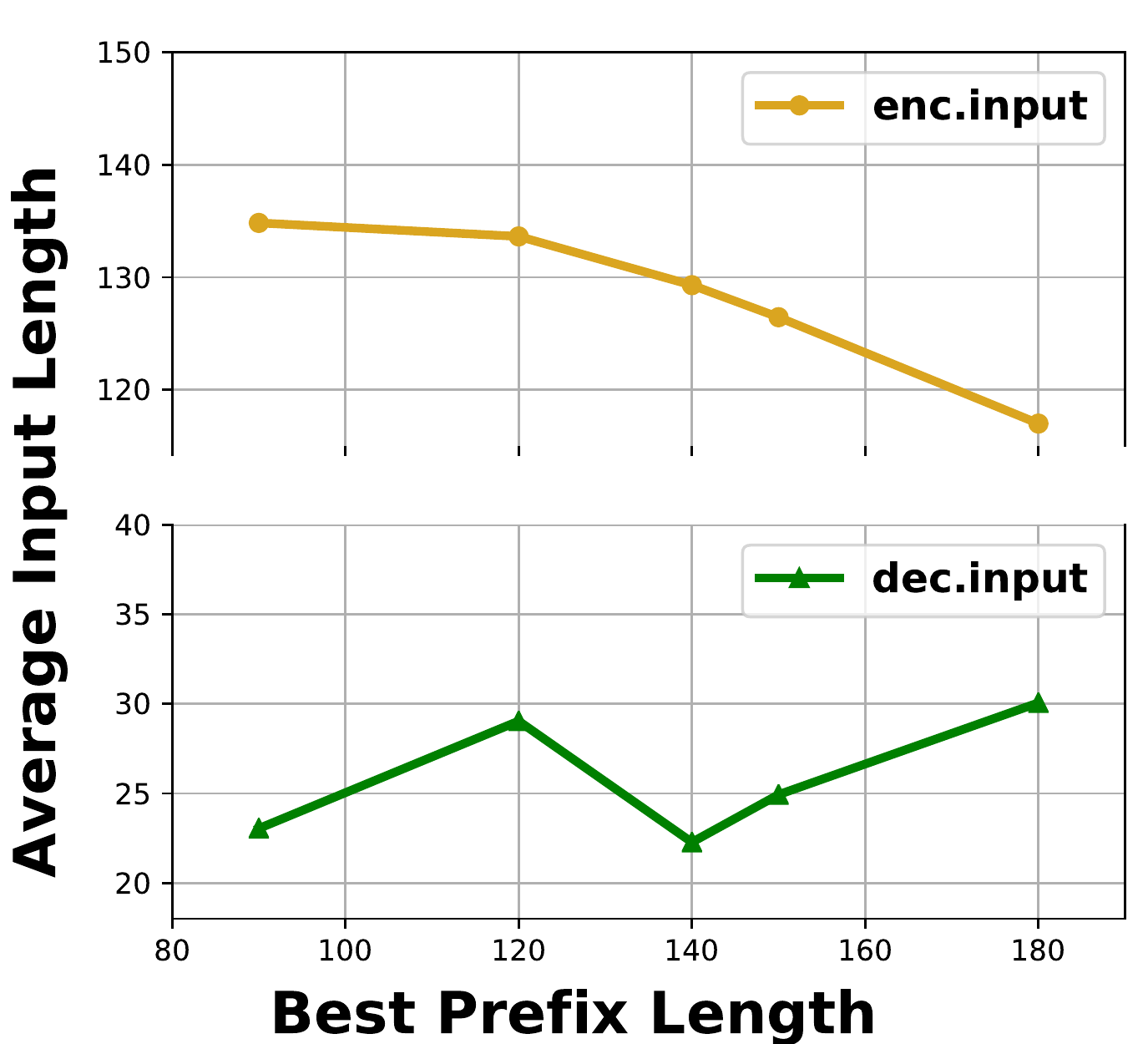}
    }
    \subfigure[QMSum]{
        \includegraphics[scale=0.25]{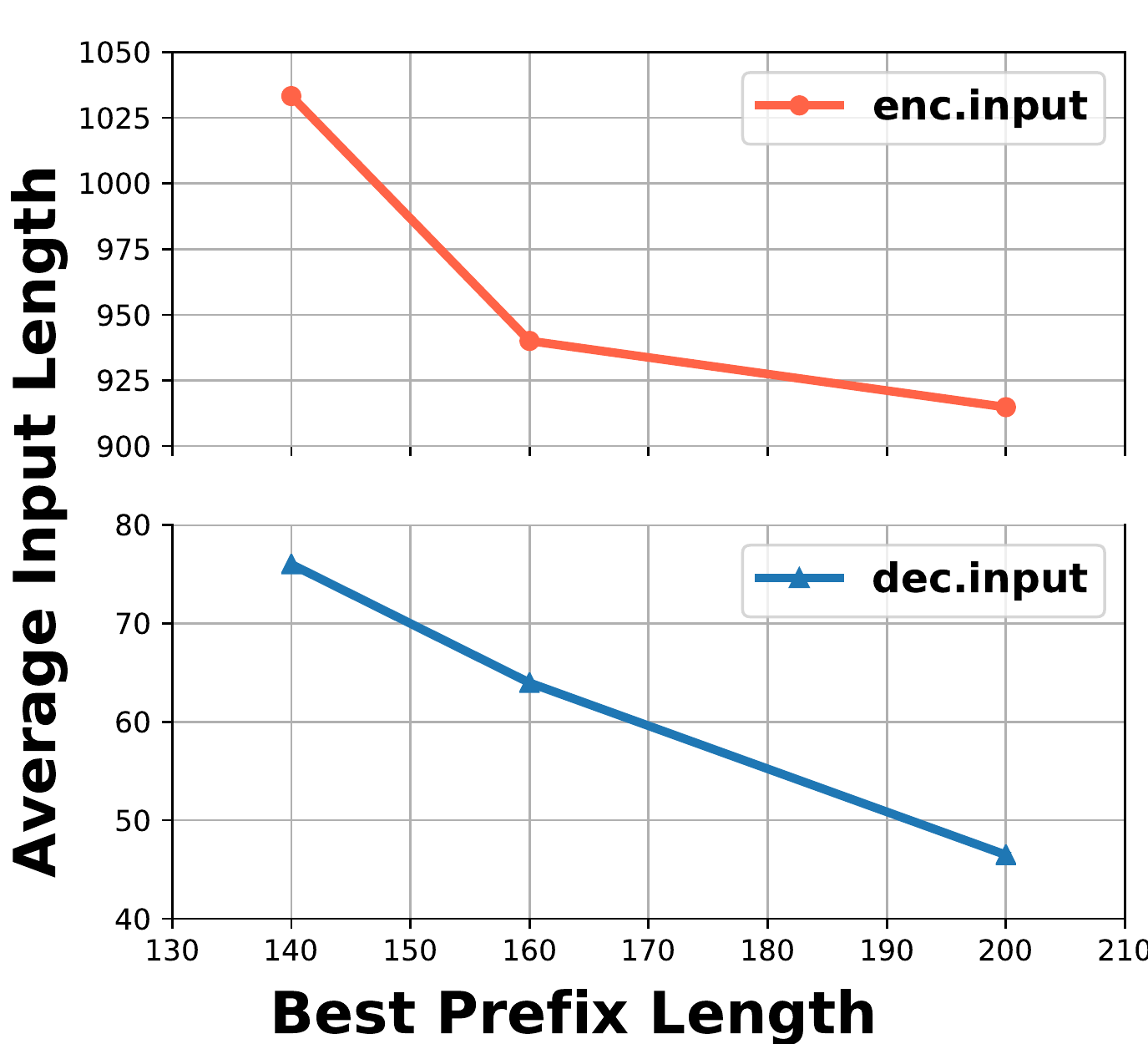}
    }
    \vspace{-0.3cm}
    \caption{Relationships between length of encoder/decoder input and the best prefix length.}
    \label{length}
    \vspace{-0.5cm}
\end{figure}

Through experiments, we explore an interesting thing, that is, the prefix length (number of domain words) that makes the model perform best may be related to the input length. Based on this assumption, we collect the source input length, target input length, and their corresponding optimal prefix length from two datasets, as shown in Figure \ref{length}. We conclude a general rule that the longer inputs may prefer the shorter prefix. This phenomenon may serve as a research point in the future.

\section{Discussion}
\subsection{Human Evaluation}
We further conduct a manual evaluation to assess the models. We randomly select 50 samples from each target domain dataset of TODSum \cite{zhao2021todsum} and ask 10 professional linguistic evaluators to score the ground truth and summaries generated by 4 models according to 5 metrics: fluency, informativeness, factual consistency, domain relevance, and redundancy. Each metric is rated by 3 evaluators from 1 (worst) to 5 (best) and the scores for each summary are averaged. Note that the intra-class agreement score is 0.605.

As shown in Table \ref{tab:human_evaluation}, the fluency scores of all models are high, which is because that abstractive models fine-tuned on contextualized language backbones can generate fluent sentences \cite{Lewis2020BARTDS}. For factual consistency, both DOP and BART w. DS achieve better performance than Prefix-tuning, which suggests that the dialogue state information guides the model to focus more on the key information, such as slots and intents. Besides, the DOP outperforms all baselines in the domain relevance metric. This demonstrates that the domain-oriented prefix module plays a crucial role in enhancing the ability of the model to identify domain-related features and disentangle the knowledge of the source and target domains. Surprisingly, the scores about the redundancy of Prefix-tuning and DOP are higher than that of BART and BART w. DS.  This is because the model can efficiently extract key contents from a limited amount of data without relying on large-scale pre-trainings.

\begin{table}[]
\centering
\resizebox{0.48\textwidth}{!}{%
\begin{tabular}{l|ccccc}
\hline
\textbf{Model} & \textbf{Flu.} & \textbf{Inf.} & \textbf{Fac.} & \textbf{Dom.} & \textbf{Red.} \\
\hline
\textbf{Ground Truth} & 4.95 & 4.56 & 4.28 & 4.71 & 4.33 \\
\textbf{BART} & 4.19 & 4.21 & 3.55 & 3.19 & 3.53\\
\textbf{BART w. DS} & 4.36 & 4.34 & 4.09 & 3.35 & 3.62 \\
\textbf{Prefix-tuning} & 4.23 & 4.29 & 3.67 & 3.28 & 4.09 \\
\textbf{DOP} & 4.68 & 4.42 & 4.10 & 4.07 & 4.13\\
\hline
\end{tabular}}
\caption{Human evaluation on Fluency (Flu.), Informativeness (Inf.), Factual Consistency (Fac.), Domain Relevance (Dom.), and Redundancy (Red.) for TODSum datatset.}
\label{tab:human_evaluation}
\vspace{-0.5cm}
\end{table}

\section{Challenges}
Through the analysis of cases in Appendix \ref{case}, we summarize two challenges of low-resource domain adaption for abstractive dialogue summarization task:

\textbf{1.\ Confusion between domains with high similarity:}  We found that in domains with high-overlap vocabularies, i.e., \emph{restaurant} and \emph{hotel}, \emph{train} and \emph{taxi}, the model generates some domain-confusing sentences. Taken \emph{hotel-restaurant} pair as an example, when \emph{restaurant} is as the target domain, a sentence like "book a restaurant room that can accommodate 3 people, ..." is generated, which is more likely to exist in the \emph{hotel} domain. Note that this challenge does not affect the accuracy of key elements, but the language style is not appropriate.

\textbf{2.\ Information dispersion:} Because of the long input sequence, it is difficult for the models to pay attention to all aspects of the long dialogue and there will be problems with attention deviations on the key elements of dialogues, especially for this lightweight and small parameter training paradigm.

\section{Conclusion}
In this paper, we present a domain-oriented prefix-tuning model to handle the domain adaptation for dialogue summarization based on an efficient and generalizable fine-tuning method. The domain word initialized prefix module disentangles the target domain knowledge from the source domain and the discrete prompts enhance the generalization ability of the model. The experiments in zero-shot and few-shot settings show that our methods have made great progress on two datasets.

\section*{Acknowledgement}
We thank all anonymous reviewers. This work was partially supported by National Key R\&D Program of China No. 2019YFF0303300 and Subject II  No. 2019YFF0303302, DOCOMO Beijing Communications Laboratories Co., Ltd, MoE-CMCC "Artificial Intelligence" Project No. MCM20190701, BUPT Excellent Ph.D. Students Foundation CX2021204.

\bibliography{anthology,custom}
\bibliographystyle{acl_natbib}

\appendix

\section{Experiment details}
\label{sec:appendix1}





\subsection{Baselines}
\label{baselines}

We describe baselines in detail as follows.

\textbf{Lead-3}: This method is commonly used in news summarization task, which treats the first three sentences of the document as the summary.

\textbf{Oracle}: The method is used to obtain an oracle through a greedy way similar to \citet{nallapati2017summarunner}, which treats the sentences that maximize the the ROUGE-2 as the summary.

\textbf{BertExt}: Proposed by \citet{liu2019text}, this model is extractive and its parameters are initialized with BERT.

\textbf{BertAbs}: Proposed by \newcite{liu2019text}, this model is abstractive, which encoder is initialized with BERT and is trained with a different optimizer than decoder.

\textbf{BART}: This model is proposed by \citet{Lewis2020BARTDS}, which is a state-of-the-art abstractive summarization model pre-trained with a denoising autoencoding objective.

\textbf{BART w. DS/QR}: This model is proposed by \citet{zhao2021todsum}, which is a general summarization framework, with with two encoders that share the underlying parameters and a decoder, which can fuse the input text and dialogue state/query.

\textbf{Prefix-tuning}: This model is proposed by \citet{li2021prefix}, which introduces a prefix matrix on the basis of fixed pre-training BART parameters and allows the prefix matrix to learn task information through training, which optimizes the summarization performance in the small parameters and few-shot scenarios.

\textbf{QMSum}: This model is proposed by \citet{zhong-etal-2021-qmsum}, which is a two-stage locate-then-summarize solution on query-based meeting summarization task.

\subsection{Evaluation Metrics}
\label{metric}

We use the ROUGE \cite{lin2004rouge,lin2004automatic}\footnote{https://pypi.org/project/rouge/} metrics to quantitatively evaluate the performance of our model. Rouge (Recall-Oriented Understudy for Gisting Evaluation) is metrics for evaluating summarization. It calculates by comparing the generated summary with references to obtain the corresponding score to measure the similarity between them.

\section{Parameter Scale of Models}
We show the amount of \textbf{trainable} parameters for our DOP model and other baseline models in Table \ref{param}. Among the full-parameter fine-tuning methods, except for the relatively simple PGN model, other models have reached the scale of hundreds of megabytes, which will take up a lot of time and space in model training and storage. Prefix-tuning and DOP-tuning greatly reduce the storage space of the model, improving the efficiency of the model. Compared with prefix-tuning, our method achieves better results with fewer parameters.

\begin{table}[]
\centering
\small
\begin{tabular}{|l|c|}
\hline
\textbf{Models}        & \textbf{Trainable Parameters} \\ \hline
\textbf{BertExt}       & 120.51M                    \\
\hline
\textbf{PGN}           & 27.64M                        \\
\textbf{Transformer}   & 257M                          \\
\textbf{BertAbs}       & 180.22M                          \\
\hline
\textbf{BART-large}    & 400M                          \\
\textbf{BART w. DS/QR} & 406M                           \\
\hline
\textbf{Prefix-tuning} & 81.82M                        \\
\textbf{DOP-tuning}    & 61.52M                        \\ \hline
\end{tabular}
\caption{Trainable parameter scales of different models, where "DS" means the dialogue state in TODSum and "QR" means the query in QMSum.}
\label{param}
\end{table}

\section{Case Study}
\label{case}
\begin{figure*}[htp]
\centering
\includegraphics[scale=0.6]{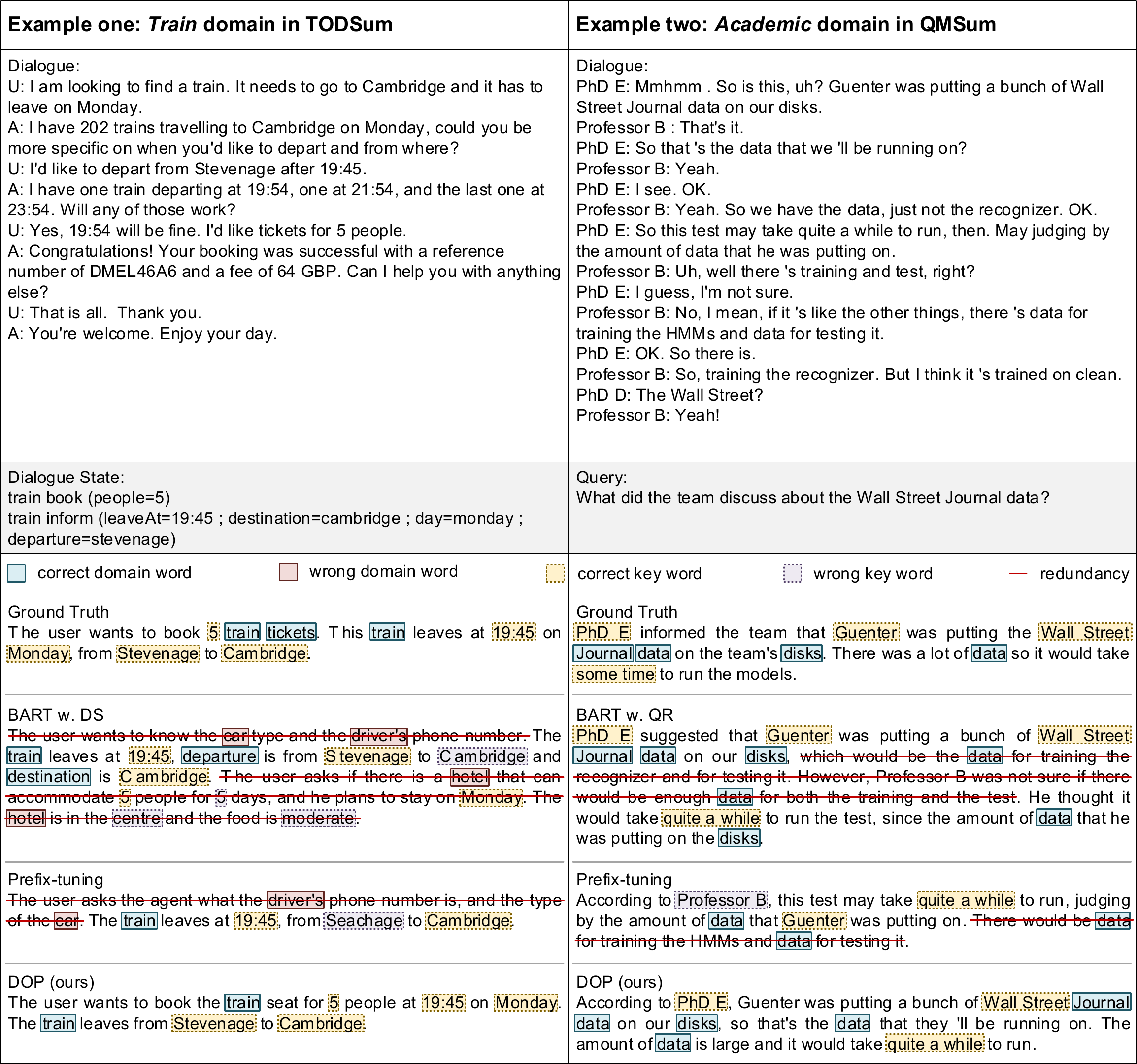}
\caption{Case study for two examples from TODSum and QMSum datasets. We present the dialogue, its corresponding dialogue state/query, ground truth, BART w. DS prediction, Prefix-tuning prediction, and prediction of our DOP model.}
\label{fig:case}
\end{figure*}

Figure \ref{fig:case} shows two examples from the TODSum and QMSum respectively. For example one of \emph{train} domain in TODSum, BART w. DS generates some incorrect and redundant information related to the \emph{taxi} domain and \emph{hotel} domain. To make matters worse, for \emph{train} domain, it loses the intent of the user about booking tickets and wrongly generates the key information, i.e., the departure location. The Prefix-tuning still confuses the knowledge of \emph{train} and \emph{taxi} domains, that is, the booking intent in the \emph{train} domain is wrongly predicted as the user wants to know something about "cars". Moreover, the quality of the generated key information is not high, i.e., the wrong departure location "Seachage" and the missing time "Monday". For example two of \emph{academic} domain in QMSum, both BART w. QR and Prefix-tuning predict too many details of the dialogue, which makes the summary redundant. Besides, Prefix-tuning generates the wrong speaker "Professor B", which leads to the summary being inconsistent. 

Compared to the above two models, our method solves some difficult issues in low-resource domain adaptation for dialogue summarization. By initializing prefix matrix with domain words, our model achieves domain disentanglement and the prediction of domain-related information is basically accurate. Through discrete prompts, our model has the ability to generalize to new domains and the accuracy of prediction about domain-independent key information is greatly improved.

\section{Domain Words}
\label{sec:domain_words}

\begin{figure*}[htp]
\centering
\includegraphics[scale=0.55]{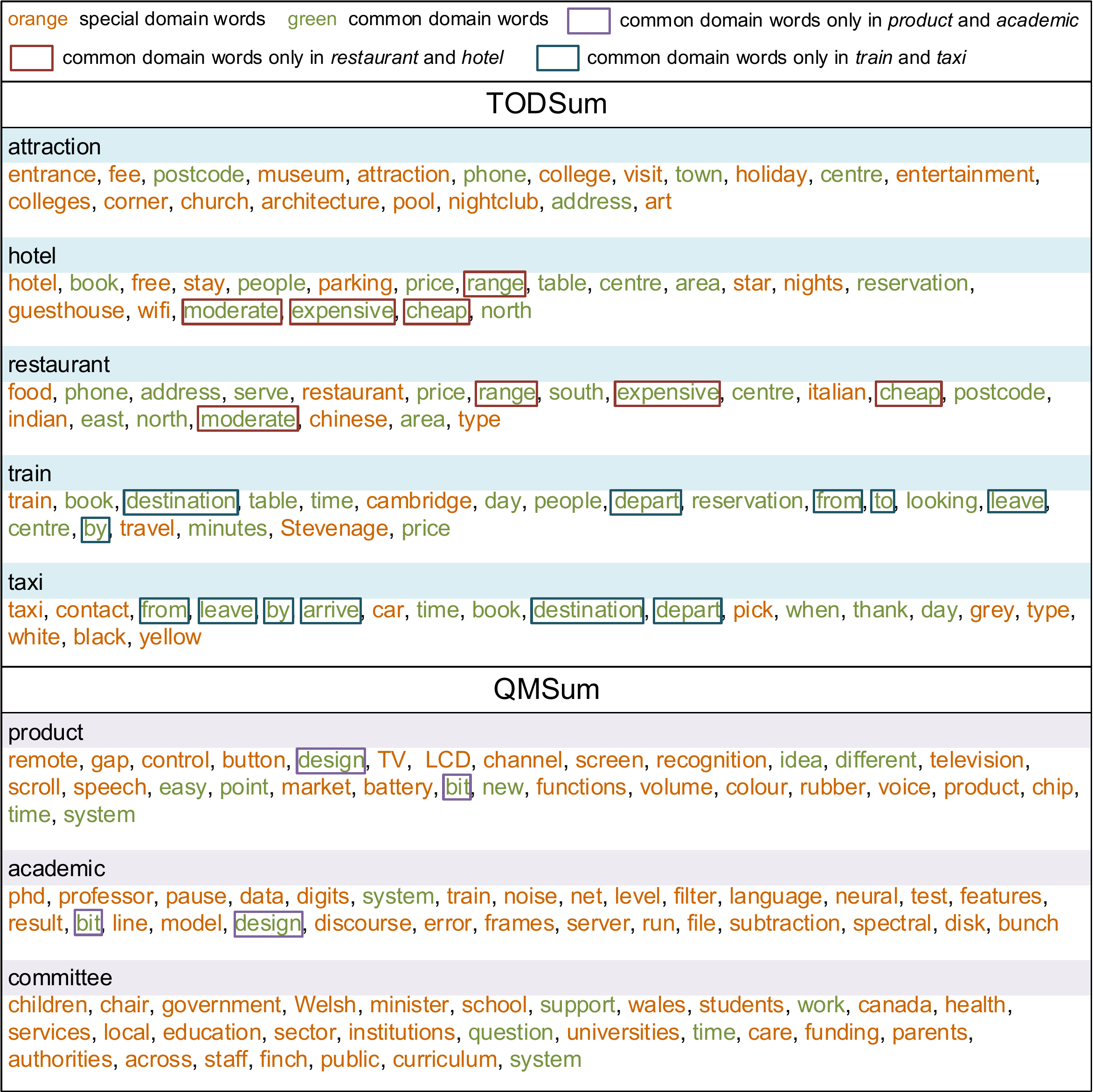}
\caption{Domain words in each domain of two datasets. We use different signs to mark the special domain words, common domain words, and common domain words in specific domains separately.}
\label{fig:domain_words}
\end{figure*}

We present some domain words for each domain in Figure \ref{fig:domain_words}. In order to facilitate reading, we only show some of the domain words, that is, we select the first 20 words for each domain of TODSum, and the first 30 words for each field of QMSum as display.

For TODSum, we can see that there are relatively many common domain words, which are more concentrated on location words, or some information, such as "phone", "postcode" and so on. In addition, there are many common domain words that only appear in \emph{restaurant} \& \emph{hotel} or \emph{train} \& \emph{taxi}. For example, price-related descriptions are usually mentioned when booking a restaurant or hotel, and "destination", "depart", "from", "to" are usually mentioned when booking train tickets or taking a taxi. Special domain words can better distinguish different domains. There will be more special domain words in \emph{attraction}, such as "entrance", "college", "nightclub", etc., which will not appear in other domains. Besides, users will mention the star rating when booking a hotel, and want to know the food type when booking a restaurant seat. When booking train tickets, they usually plan to travel, and when taking a taxi, they want to know the color of the car.

Compared with TODSum, QMSum has many more special field words, because the three fields contained in QMSum are more different. For \emph{product}, participants will discuss various features of products such as TVs, LCDs, etc., such as screens, buttons, colors, and functions. For \emph{academic}, participants generally discuss models, experimental data, or some errors. And for \emph{committee}, participants generally discuss student education or national government issues. Common domain words have only some generral words, such as "different", "system", etc., and only a few special common domain words exist in \emph{product} \& \emph{academic}, such as "design", "bit" and other technology-related words.

\end{document}